\documentclass{article}

\usepackage{arxiv}

\usepackage[utf8]{inputenc} % allow utf-8 input
\usepackage[T1]{fontenc}    % use 8-bit T1 fonts
\usepackage{hyperref}       % hyperlinks
\usepackage{url}            % simple URL typesetting
\usepackage{booktabs}       % professional-quality tables
\usepackage{amsfonts}       % blackboard math symbols
\usepackage{nicefrac}       % compact symbols for 1/2, etc.
\usepackage{microtype}      % microtypography
\usepackage{lipsum}		% Can be removed after putting your text content
\usepackage{graphicx}
\usepackage{natbib}
\usepackage{doi}
\usepackage{here}     
\usepackage{emptypage}
\usepackage{subcaption}
\usepackage{multicol}
\usepackage[dvipsnames]{xcolor}

\usepackage{amssymb}
\usepackage{amsmath}
\usepackage{amsthm}
\usepackage{bm}

\usepackage{algpseudocode}
\usepackage{algorithm}

\graphicspath{ {./img/} }  % where images are stored

\newcommand{\set}[1]{\ensuremath{\left\{{#1}\right\}}}

\newcommand{\RR}{\ensuremath{\mathbb{R}}}

\newcommand{\TTT}{\ensuremath{\mathcal{T}}}

\newcommand{\WWW}{\ensuremath{\mathcal{W}}}
\newcommand{\DDD}{\ensuremath{\mathcal{D}}}
\newcommand{\III}{\ensuremath{\mathcal{I}}}
\newcommand{\LLL}{\ensuremath{\mathcal{L}}}

\newcommand{\mb}[1]{\mathbf{#1}}

\newcommand{\titleString}{
Writer adaptation for offline text recognition: An exploration of neural network-based methods
}

\title{\titleString}

\author{
	{Tobias van der Werff$^*$}\\
	Department of Artificial Intelligence\\
	University of Groningen\\
	9747 AG Groningen, The Netherlands \\
	\texttt{t.n.van.der.werff@rug.nl} \\
	\And
	\href{https://orcid.org/0000-0002-7548-3858}{\includegraphics[scale=0.06]{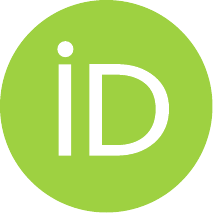}\hspace{1mm}Maruf A. Dhali} \\
	Department of Artificial Intelligence\\
	University of Groningen\\
	9747 AG Groningen, The Netherlands \\
	\texttt{m.a.dhali@rug.nl} \\
        \And 
        \href{https://orcid.org/0000-0003-2351-930X}{\includegraphics[scale=0.06]{orcid.pdf}\hspace{1mm} Lambert Schomaker}\\
	Department of Artificial Intelligence\\
	University of Groningen\\
	9747 AG Groningen, The Netherlands \\
	\texttt{l.r.b.schomaker@rug.nl} \\
}

\renewcommand{\headeright}{Preprint}
\renewcommand{\undertitle}{Preprint}
\renewcommand{\shorttitle}{Writer adaptation for offline text recognition}

\hypersetup{
pdftitle={\titleString},
pdfauthor={Tobias van der Werff, Maruf A. Dhali, Lambert Schomaker},
pdfkeywords={Offline handwritten text recognition, Writer adaptation, Few-shot adaptation, Conditionality},
}

\begin{document}
\maketitle

%%%%%%%%%%%%%%%%%%%%%%%%%%%%%%%%%%%%%%%%%%%%%%%%%%%%%%%%%%%%%%%%%%%%%%%%%%%%%%%%%%%
%%%%%%%%%%%%%%%%%%%%%%%%%%%%%%%%%%%%%%%%%%%%%%%%%%%%%%%%%%%%%%%%%%%%%%%%%%%%%%%%%%%
\begin{abstract}

Handwriting recognition has seen significant success with the use of deep learning. However, a persistent shortcoming of neural networks is that they are not well-equipped to deal with shifting data distributions. In the field of handwritten text recognition (HTR), this shows itself in poor recognition accuracy for writers that are not similar to those seen during training. An ideal HTR model should be adaptive to new writing styles in order to handle the vast amount of possible writing styles. In this paper, we explore how HTR models can be made writer adaptive by using only a handful of examples from a new writer (e.g., 16 examples) for adaptation. Two HTR architectures are used as base models, using a ResNet backbone along with either an LSTM or Transformer sequence decoder. Using these base models, two methods are considered to make them writer adaptive: 1) model-agnostic meta-learning (MAML), an algorithm commonly used for tasks such as few-shot classification, and 2) writer codes, an idea originating from automatic speech recognition. Results show that an HTR-specific version of MAML known as MetaHTR improves performance compared to the baseline with a 1.4 to 2.0 improvement in word error rate (WER). The improvement due to writer adaptation is between 0.2 and 0.7 WER, where a deeper model seems to lend itself better to adaptation using MetaHTR than a shallower model. However, applying MetaHTR to larger HTR models or sentence-level HTR may become prohibitive due to its high computational and memory requirements. Lastly, writer codes based on learned features or Hinge statistical features did not lead to improved recognition performance. \footnote{Code used for this research can be found at \url{https://github.com/tobiasvanderwerff/master-thesis} \\
$^*$ Corresponding author}

\end{abstract}


%%%%%%%%%%%%%%%%%%%%%%%%%%%%%%%%%%%%%%%%%%%%%%%%%%%%%%%%%%%%%%%%%%%%%%%%%%%%%%%%%%%
%%%%%%%%%%%%%%%%%%%%%%%%%%%%%%%%%%%%%%%%%%%%%%%%%%%%%%%%%%%%%%%%%%%%%%%%%%%%%%%%%%%
\section{Introduction}

Handwriting recognition has seen major successes using deep learning, manifested in domains like handwritten text recognition~\citep{michael2019evaluating,ameryan2021limited}, writer identification~\citep{yang2016deepwriterid,he2020fragnet}, binarization~\citep{dhali2019binet}, and word spotting~\citep{chanda2018zero}. However, neural networks are often still lacking when it comes to adapting to novel environments~\citep{kouw2019review}. Arguably, much of the modern success of deep learning can be attributed to collecting massive amounts of data to cover as many parts of the underlying data distribution as possible, combined with a proportional increase in computing power and model size~\citep{kaplan2020scaling}. However, such a brute-force approach to learning is often not practical for handwriting recognition tasks. Large, high-quality corpora of annotated handwritten texts are often scarce, especially for historical handwriting. In this case, more efficient use of data and reusability of previously learned representations becomes important.

\begin{figure}
    \centering
    \includegraphics[width=0.9\textwidth]{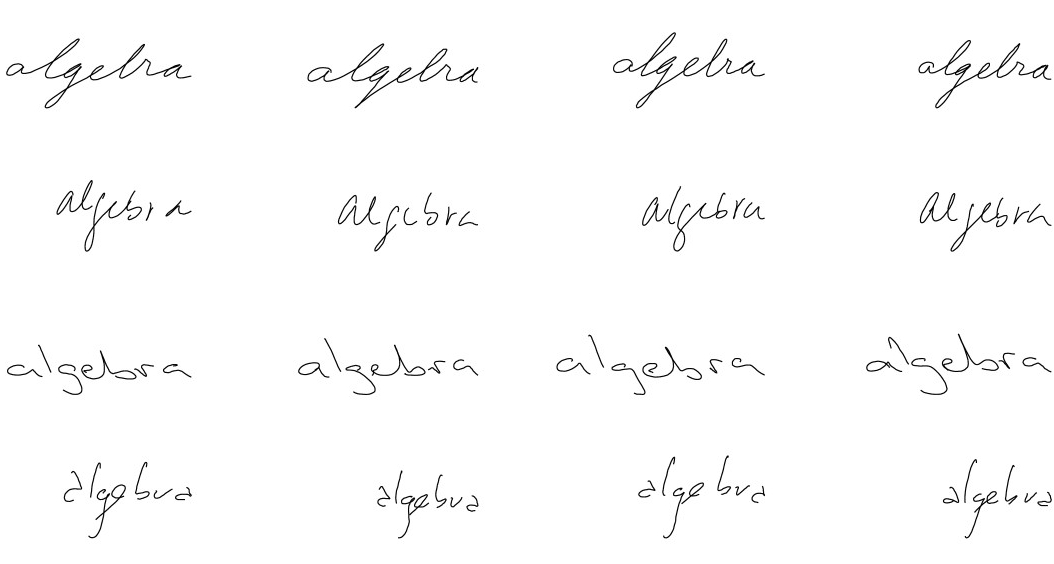}
    \caption{The word ``algebra'' written by different writers. Each row contains handwriting for a single writer, recorded at four different times. Note that variation manifests itself between writers but also within individual writers. Figure taken from~\citet{305ba571ed2f4e7ea01992ae48e26ac9}.}
    \label{fig:writer-variation-and-variability}
\end{figure}

In this paper, we focus on improving one of the most common handwriting recognition tasks: handwritten text recognition (HTR), which refers to the process of automatically turning images of handwritten text into letter codes. HTR remains a challenging problem, mainly due to the large number of possible handwriting variations (Fig.~\ref{fig:writer-variation-and-variability}). In this research, we attempt to make modern HTR models \textit{writer adaptive}, referring to the idea that when a trained HTR model is presented with a novel writing style, it is able to modify its internal representations in such a way as to improve recognition performance for that style. We focus on cases with limited data available for adaptation (10-20 samples), as this represents a realistic scenario for real-time adaptation. In a practical setting, a user of an HTR system could be asked to supply a handful of handwriting examples in order to improve recognition performance on their writing style. How to perform writer-specific adaptation effectively remains an open problem. A popular approach for adapting existing deep learning models is \textit{transfer learning}, where previously learned model parameters are reused for a new but related task that has only a modest amount of training data, leading to notable successes in fields such as natural language processing~\citep{devlin2018bert} and computer vision~\citep{oquab2014learning}.

It is important to note that the potential benefit of including writer identity as a conditional variable cannot easily be decoupled from architectural choice. For example, Hidden Markov Models~\citep{baum1966hmm} have been a common choice for HTR in the past, and methods have been developed to include writer identity in such models. However, these methods are often not usable for modern approaches to HTR using deep neural networks, which use powerful hierarchical representations that outperform past methods. In this sense, a relevant question is whether state-of-the-art deep learning approaches to HTR can benefit from explicit writer information \textit{in the first place}. We will show that this benefit is not obvious, providing at best modest improvements compared to a writer-unaware baseline.

There are several problems at hand. In order to adapt effectively based on style information, there is a clear need to identify \emph{what exactly a deep learning model has not learned yet}. The question can be formulated as ``What novelty does this new writer introduce that is not effectively handled by the neural network?''. Another relevant question is what signal source can be provided to allow for adaptation, and the non-trivial question of effectively including such information into an HTR model. We draw inspiration from a recently published paper by~\citet{bhunia2021metahtr} which employs meta-learning to flexibly adapt HTR models to different writers, seemingly with great success. Meta-learning (also known as learning-to-learn) is currently an active area of research~\citep{hospedales2020metasurvey}.

Meta-learning is concerned with improving the learning algorithm itself. Often, the idea is to adapt a learning algorithm to a new task based on a small number of task-specific examples. The aim is to learn underlying \textit{meta-knowledge} that can be transferred to various tasks, even those unseen during training. The paper by~\citet{bhunia2021metahtr} makes use of a modified form of model-agnostic meta-learning (MAML)~\citep{finn2017maml}, which they call MetaHTR. As this is one of the more promising ideas for writer-aware adaptation, we explore several versions of the MAML approach and will test its ability to perform writer-specific adaptation.

Additionally, we experiment with another approach, based on \textit{writer codes}: Compact vector representations of individual writers that are supposed to capture the most relevant information about a writer to allow for effective adaptation. Writer codes can be learned or explicitly given as part of the model input. The codes are inserted into a trained HTR model by adjusting the parameters of batch normalization layers. We experiment with several approaches to creating such a writer code: One based on learned feature vectors and one based on traditional handcrafted features used for writer identification. Although this approach is conceptually appealing, our version of writer codes does not yield concrete benefits for adaptation.

We summarize the contributions in this paper as follows:

\begin{itemize}
    \item We show that MAML-based methods applied to a trained HTR model can lead to
        improved data efficiency, showing an improvement between 1.4 and 2.0 word
        error rate compared to a naive fine-tuning baseline;

    \item We test the capability of MetaHTR to perform writer-specific adaptation, finding
        that it leads to an improvement of 0.7 word error rate for a deep HTR model,
        but shows no significant effect for smaller models;

    \item We analyze how a trained HTR model can be effectively adapted based on
        writer-specific vector representations, finding that fine-tuning batch
        normalization scale and bias parameters can be an effective way to obtain
        additional performance gains, even without writer-specific information;

    \item We show that writer codes based on learned features or Hinge statistical features do not lead to improved recognition performance.
\end{itemize}

This paper is structured as follows. In Section~\ref{chap:related-work}, we provide related works. In Section~\ref{chap:methods}, we propose several techniques for writer-adaptive HTR and experiments to verify their performance. In Section~\ref{chap:experiments}, we outline our experimental setup. In Section~\ref{chap:results}, we show results for the proposed methods, and finally, in Section~\ref{chap:discussion-and-conclusion} and Section~\ref{chap:conclusion}, we discuss the results and future work.

%%%%%%%%%%%%%%%%%%%%%%%%%%%%%%%%%%%%%%%%%%%%%%%%%%%%%%%%%%%%%%%%%%%%%%%%%%%%%%%%%%%
%%%%%%%%%%%%%%%%%%%%%%%%%%%%%%%%%%%%%%%%%%%%%%%%%%%%%%%%%%%%%%%%%%%%%%%%%%%%%%%%%%%
\section{Related works} \label{chap:related-work}

\paragraph{Handwritten text recognition:}
Early approaches to HTR often employed Hidden Markov Models~\citep{bianne2011hmm} (HMM). More recently, the field of HTR has progressed from HMM-based methods to end-to-end trainable neural networks with many layers. Recurrent neural networks (RNN), and in particular Multi-dimensional Long Short-Term Memory (MDLSTM), networks~\citep{graves2007multi} have been commonly used sequence modeling architectures for HTR models~\citep{puigcerver2017htr}. The MDLSTM architecture, in combination with the Connectionist Temporal Classification~\citep{graves2006ctc} loss (CTC), served as a replacement for Hidden Markov Model-based methods~\citep{graves2008offline}. Whereas standard RNN architectures process data along a one-dimensional axis -- e.g., a time axis --, the MDLSTM architecture allows recurrence across multi-dimensional sequences, such as images. In more recent years, it has been observed that the expensive recurrence of the MDLSTM could be replaced by a CNN + bidirectional LSTM architecture~\citep{shi2016endtoendhtr,puigcerver2017htr}. The CNN-RNN hybrid + CTC has been a commonly used architecture (e.g.,~\citet{dutta2018cnnrnn, sueiras2018offline, wigington2017data}). For example, in~\citet{dutta2018cnnrnn}, a spatial transformer network, residual convolutional blocks (ResNet-18), stacked BiLSTMs, and a CTC layer are used. 

Although CTC has been a common decoding method, some of its downsides -- such as the inability to consider linguistic dependency across tokens -- have led to architectures that replace CTC in favor of attention modules~\citep{bahdanau2014attention}. Attention-based encoder-decoder architectures have reached state-of-the-art performance in recent years~\citep{michael2019evaluating}. Attention alleviates constraints on input image sizes and the need for segmentation or image rectification~\citep{jaderberg2015stn} for irregular images. This thus allows for simplification in the design of HTR architectures. In~\citet{li2019show}, a ResNet-31 is combined with an LSTM-based encoder-decoder along with a 2-dimensional attention module for irregular text recognition in natural scene images.

A trend in recent years has been to replace the linear recurrence of RNNs with the more parallelizable Transformer architecture and attention-based approaches more broadly. In a recent work~\citep{diaz2021rethinking}, various architectures for universal text line recognition are studied, using various encoder and decoder families. The authors find that a CNN backbone for extracting visual features, coupled with a Transformer encoder, a CTC decoder, and an explicit language model, is the most effective approach for recognizing line strips. Building on top of the idea~\citep{dosovitskiy2020vit} of using Transformer-only architectures for vision tasks,~\citet{li2021trocr} explore an end-to-end Transformer encoder-decoder architecture for text recognition, initialized with a pretrained vision Transformer for extracting visual features and a pretrained RoBERTa~\citep{liu2019roberta} Transformer for sequence decoding. After initialization, the model is pretrained on large-scale synthetic handwritten images and fine-tuned on a human-labeled dataset.

%%%%%%%%%%%%%%%%%%%%%%%%%%%%%%%%%%%%%%%%%%%%%%%%%%%%%%%%%%%%%%%%%%%%%%%%%%%%%%%%%%%%
\paragraph{Meta-learning:}
Meta-learning, or learning-to-learn, is an alternative paradigm to traditional neural network training, which aims to improve the learning algorithm itself~\citep{hospedales2020metasurvey}. By learning shared knowledge across various tasks over multiple learning episodes, the aim is to improve future learning performance. The main meta-learning method we focus on here is Model-Agnostic Meta-Learning~\citep{finn2017maml} (MAML). MAML aims to find a parameter initialization such that a small number of gradient updates using a handful of labeled samples produces a classifier that works well on validation data. MAML is related to transfer learning, in the sense that finding good initialization parameters for a model to facilitate adaptation to various tasks plays a central role. Due to its model-agnostic nature, MAML can be applied to various application domains without significant modifications.

Due to the inner/outer-loop optimization process, MAML has great flexibility in terms of the kinds of parameters that can be learned in the inner loop, e.g., parameterized loss functions~\citep{bechtle2021metaloss}, learning rates~\citep{li2017metasgd}, and attenuation weights~\citep{baik2020l2f}.  Meta-learning has been applied to various areas such as reinforcement learning and few-shot classification, but, notably, also to speech recognition, in the form of accent adaptation~\citep{winata2020learning} and speaker adaptation~\citep{klejch2018learning}. MetaSGD~\citep{li2017metasgd} is a modification of MAML and involves learning the update direction and learning rate along with the parameter initialization. MAML++~\citep{antoniou2018mamlplusplus} addresses the training instability of MAML that is commonly observed. MAML has also been used in combination with other types of meta-learning. For example, in~\citet{rusu2018leo}, the authors combine MAML with model-based meta-learning, using a latent generative representation of model parameters and applying MAML in this lower-dimensional latent space.

%%%%%%%%%%%%%%%%%%%%%%%%%%%%%%%%%%%%%%%%%%%%%%%%%%%%%%%%%%%%%%%%%%%%%%%%%%%%%%%%%%%%
\paragraph{Writer adaptation:}
Many early approaches for writer adaptation are proposed for HMMs using Gaussian Mixture Models. For example,~\citet{vinciarelli2002writer} use linear transformations between original parameters and re-estimated parameters for adjusting GMM parameters using maximum likelihood linear regression. More recently, there have been several attempts at adaptation in the space of HTR using neural networks. In~\citet{nair2018knowledge}, the authors perform simple fine-tuning on a new handwriting collection, showing that this can lead to efficient transfer between datasets using a limited amount of fine-tuning data. In~\citet{szummer2006discriminative}, the authors cluster writers by style and train a classifier for each cluster, using a mixture-of-experts setup for choosing the best combination of classifiers. For a new writer, the combination of classifiers is based on classification confidence for that writer.  In~\citet{zhang2012writer}, the authors learn a linear writer-specific feature transformation in order to create a style-invariant classifier, which they call Style Transfer Mapping (STM).  Whereas the original approach was not used in the context of neural networks, a later approach~\citep{zhang2017online} uses STM for neural networks in the context of Chinese character recognition. In~\citet{wang2020writer}, the authors employ writer codes for writer-specific Chinese handwritten text recognition using a CNN-HMM hybrid model. They feed a writer code into adaptation layers tied to individual convolution layers. The result is added element-wise to the intermediate CNN feature maps. At train time, writer codes are jointly learned with the adaptation layers. At test time, codes for new writers are randomly initialized and optimized using one to three gradient steps. Recently,~\citet{wang2022fast} used a style extractor network trained on a writer identification task to extract a writer code, used to adapt a writer-independent recognizer. Specifically, the writer code is added to the convolutional layer output after being fed through a fully-connected layer. 

The writer adaptation problem has also been formulated as a domain adaptation problem~\citep{zhang2019sequence, kang2020unsupervised, yang2018deep}. In~\citet{zhang2019sequence}, a gated attention similarity unit is used to find character-level writer-invariant features. In~\citet{kang2020unsupervised}, the authors employ an adversarial learning approach using synthetic data. A generic HTR model is initially trained using synthetic data and adapted to new writers using a domain discriminator network.

%%%%%%%%%%%%%%%%%%%%%%%%%%%%%%%%%%%%%%%%%%%%%%%%%%%%%%%%%%%%%%%%%%%%%%%%%%%%%%%%%%%
%%%%%%%%%%%%%%%%%%%%%%%%%%%%%%%%%%%%%%%%%%%%%%%%%%%%%%%%%%%%%%%%%%%%%%%%%%%%%%%%%%%
\section{Methodology} \label{chap:methods}

%%%%%%%%%%%%%%%%%%%%%%%%%%%%%%%%%%%%%%%%%%%%%%%%%%%%%%%%%%%%%%%%%%%%%%%%%%%%%%
\paragraph{Overview:} \label{sec:offline-htr}
An HTR model $f_{\theta}$ -- corresponding to a deep neural network --, is trained to maximize the probability $p(Y|\III; \theta)$ of the correct transcription given an input image $\III$ and ground truth character sequence $Y = (y_1, y_2,\dots,y_{L})$, where each $y_i$ is picked from a vocabulary $V$ (e.g., ASCII characters). A training dataset $\DDD = \set{(\III_1, Y_1), (\III_2, Y_2), \dots, (\III_N, Y_N)}$ consists of tuples containing an image $\mathcal{I}_i$ and the corresponding character sequence $Y_i$. The cost function is derived from cross-entropy, which, for a single example, is of the following form:

\begin{equation} \label{eq:cross-entropy-loss}
    \mathcal{L}(\mathcal{I}, Y; \theta) =
    - \frac{1}{L} \sum_{t=1}^{L} \log p(Y_t = y_t|y_{<t}, \III; \theta).
\end{equation}

%%%%%%%%%%%%%%%%%%%%%%%%%%%%%%%%%%%%%%%%%%%%%%%%%%%%%%%%%%%%%%%%%%%%%%%%%%%%%%
\subsection{Base models} \label{sec:base-models}

\begin{figure}
    \centering
    \includegraphics[trim=0 400 0 140, clip, width=0.9\textwidth]{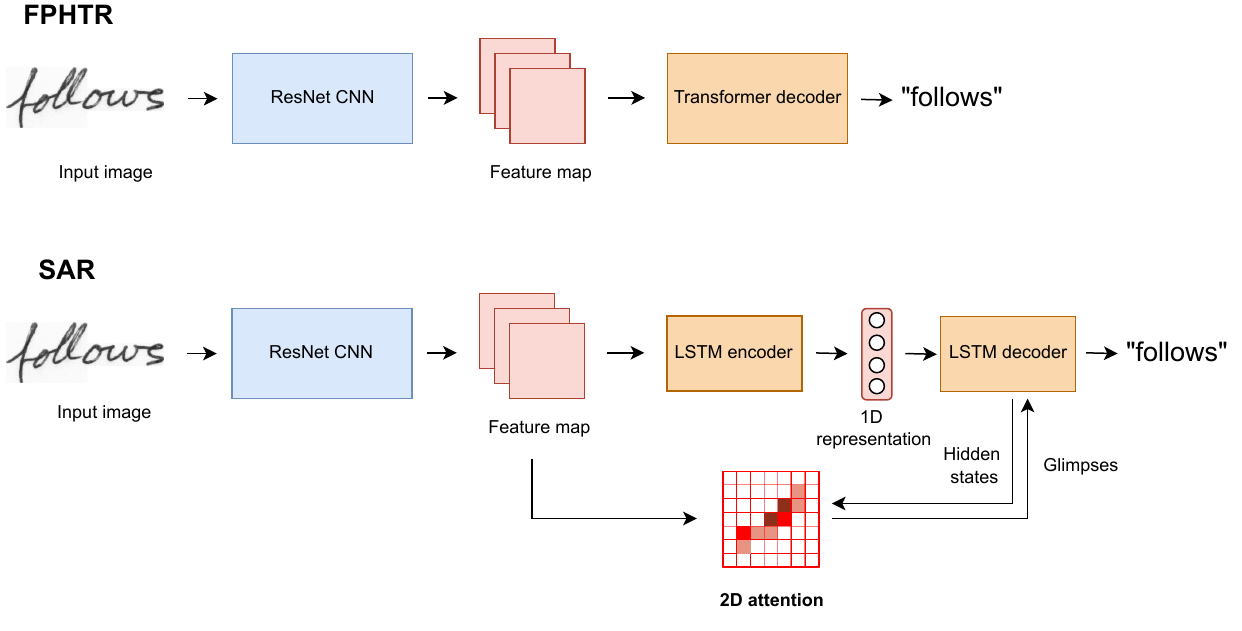}
    \caption{Schematic overview of the two base models: FPHTR and SAR.}
    \label{fig:base-models}
\end{figure}

We make use of two base models: FPHTR~\citep{singh2021fphtr} and SAR~\citep{li2019show}. FPHTR builds on the Transformer architecture, and SAR on the LSTM architecture. In Fig.~\ref{fig:base-models}, we show a high-level overview of both models to highlight their overall structure and similarity. For both models, we use two versions: a smaller version using an 18-layer ResNet backbone and a larger version with a 31-layer ResNet backbone (see Appendix~\ref{tab:base-parameters} for parameter counts). The base models are standard HTR models that do not make use of explicit writer information, chosen based on their competitive performance on common benchmarks. Their performance serves as a baseline for ``writer-unaware'' HTR models.

\subsubsection{SAR} \label{sec:lstm-based-model}

The SAR model~\citep{li2019show} is based on the Long Short-Term Memory (LSTM) architecture~\citep{hochreiter1997lstm}. It consists of a ResNet image processing backbone, LSTM encoder, LSTM decoder, and a 2-dimensional attention module. The CNN backbone consists of a modified ResNet~\citep{he2016resnet, shi2016endtoendhtr}, which outputs a 2-dimensional feature map $\mb{V}$. This is used by the consecutive LSTM encoder to extract a holistic feature vector for the whole image and also serves as context for the 2D attention network. The final encoder hidden state $\mb{h}_W$ is fed as the initial input to the LSTM decoder. A special start-of-sequence token (\textsc{\textless SOS\textgreater}) is fed as input to the decoder. At each timestep of the LSTM, a new character is sampled autoregressively. Each input at the timesteps that follow is either 1) the previous character from the ground truth character sequence (also known as \textit{teacher forcing}), or 2) the sampled character from the previous timestep (at test time). If the latter is the case, the end of the sampling procedure is signified by sampling a special end-of-sequence token (\textsc{\textless EOS\textgreater}). All token inputs are fed in as vector representations, followed by a linear transformation, $\psi(.)$. After being fed through an LSTM cell along with the previous hidden state, the timestep prediction is then calculated as $\bm{y}_t = \phi(\bm{h}'_t, \bm{g}_t) = \text{softmax}(\bm{W}_o[\bm{h}'_t; \bm{g}_t])$, where $\bm{h}'_t$ is the current hidden state and $\bm{g}_t$ is the output of the attention module. $\bm{W}_o$ is a linear transformation, which maps the features to a vector whose size is equal to the number of character classes. The attention module is a modification of the standard 1D attention module for dealing with a 2D spatial layout. It takes into account neighborhood information in the 2D plane:
\[
\begin{cases}
    \mb{e}_{ij} &= \text{tanh}(\mb{W}_v \mb{v}_{ij} + \sum_{p,q\in \mathcal{N}_{ij}}
        \mb{\tilde{W}}_{p-i,q-j} \cdot \mb{v}_{pq} + \mb{W}_h \mb{h}'_t) \\
    \alpha_{ij} &= \text{softmax}(\mb{w}_e^T \cdot \mb{e}_{ij}) \\
    \mb{g}_t &= \sum_{i, j} \alpha_{ij} \mb{v}_{ij}, \quad i=1,\dots,H, \quad j=1,\dots,W
\end{cases}
\]

Explanation of the symbols: $\mb{v}_{ij}$ is the local feature vector at position $(i, j)$ in $\mb{V}$; $\mathcal{N}_{ij}$ is the eight-neighborhood around this position; $\mb{W}_v, \mb{W}_h, \mb{\tilde{W}}$ are learned linear transformations; $\alpha_{ij}$ is the attention weight at location $(i, j)$; and $\mb{g}_t$ is the weighted sum of local features, also known as a \textit{glimpse}. The difference with a traditional attention module is the addition of the $\sum_{p,q\in \mathcal{N}_{ij}} \mb{\tilde{W}}_{p-i,q-j} \cdot \mb{v}_{pq} $ term when weighing $\mb{v}_{ij}$.

\subsubsection{FPHTR} \label{sec:transformer-based-model}

FPHTR~\citep{singh2021fphtr} is a Transformer-based architecture, consisting of a CNN backbone combined with a Transformer~\citep{vaswani2017attention} module for decoding the visual feature map into a character sequence. The architecture was originally proposed for full-document HTR, but due to its generic nature, it can easily be applied to both word and line images without any real modifications. The CNN takes an image as input and produces a 2D feature map with hidden size $d_{model}$ as output. A 2D position encoding based on sinusoidal functions is added, and the feature map is flattened into a 1D sequence of feature vectors -- each representing a position in the image --, that can be processed by the Transformer decoder. The Transformer decoder is a standard Transformer architecture~\citep{vaswani2017attention} with non-causal attention to the encoder output (it can attend to the entire output of the encoder) and causal self-attention (it can only attend to past positions of its character input). Input vectors are enhanced with 1D position encodings. Sampling is done autoregressively, in the same way as the SAR model.

%%%%%%%%%%%%%%%%%%%%%%%%%%%%%%%%%%%%%%%%%%%%%%%%%%%%%%%%%%%%%%%%%%%%%%%%%%%%%%
\subsection{Meta-learning}
Our first attempt to make HTR models writer adaptive involves meta-learning~\citep{hospedales2020metasurvey}. Adaptation occurs by providing the model with labeled examples of a writer that it should adapt to, after which the weights of the model are updated using the model-agnostic meta-learning algorithm. We first provide a brief overview of model-agnostic meta-learning in Section~\ref{sec:model-agnostic-meta-learning}, then turn to the MetaHTR approach in Section~\ref{sec:metahtr}. The explanation of these methods will be brief; for a more detailed explanation, we refer the reader to the original papers.

%%%%%%%%%%%%%%%%%%%%%%%%%%%%%%%%%%%%%%%%%%%%%%%%%%%%%%%%%%%%%%%%%%%%%%%%%%%%%%
\subsubsection{Model-agnostic meta-learning} \label{sec:model-agnostic-meta-learning}
Model-agnostic meta-learning (MAML)~\citep{finn2017maml} is an approach to meta-learning aimed at finding initial parameters that facilitate rapid adaptation. Let $p(\TTT)$ be a distribution over tasks to which a model should be able to adapt. During meta-training, a batch of tasks $\TTT_i \sim p(\TTT)$ is sampled, where samples from each task are split up in a support set $D^{tr}$ of size $K$ for adaptation (where typically $K$ is relatively small, e.g., $K \leq 16$), and a query set $D^{val}$ for testing the task-specific performance after adaptation. Training is done using stochastic gradient descent (SGD), where the model parameters $\theta$ are adapted to a task as follows:

\begin{equation} \label{eq:maml-inner-loop}
    \theta_i' = \theta - \alpha \nabla_{\theta} \mathcal{L}^{inner}(D_i^{tr}; \theta).
\end{equation}

This is referred to as the \textit{inner loop}, using an inner loop learning rate $\alpha$. After inner loop adaptation, the adapted parameters $\theta_i^{'}$ are evaluated on the query set, and the original parameters are updated by aggregating the loss over the sampled tasks, using an \textit{outer loop} learning rate $\beta$:

\begin{equation} \label{eq:maml-outer-loop}
    \theta \leftarrow \theta - \beta \nabla_{\theta} \sum_{\TTT_i \sim
    p(\TTT)}\mathcal{L}^{outer}(D^{val}_{i}; \theta_i').
\end{equation}

Whereas the inner loop optimizes for task-specific performance, the outer loop optimizes for a parameter set $\theta$ so that the task-specific training is more efficient, aiming to achieve good generalization across various tasks.

%%%%%%%%%%%%%%%%%%%%%%%%%%%%%%%%%%%%%%%%%%%%%%%%%%%%%%%%%%%%%%%%%%%%%%%%%%%%%%
\subsubsection{MetaHTR} \label{sec:metahtr}
MetaHTR is a modification of the MAML algorithm optimized for text recognition. Within the MetaHTR framework, each task instance $\mathcal{T}_i$ corresponds to a different writer. The full training process is summarized in Algorithm~\ref{alg:metahtr-train}. Once MetaHTR is trained, it can be used to rapidly adapt to specific writers at inference time. This is shown in Algorithm~\ref{alg:metahtr-inference}. With respect to MAML, MetaHTR introduces two modifications: \textit{character instance-specific weights}, and \textit{learnable layer-wise learning rates}.

\paragraph{Character instance-specific weights:} \label{sec:instance-weights}
Instance-specific weight values are added to the inner loop loss such that the model can adapt better with respect to characters having a high discrepancy. Given a ground truth character sequence $Y = \{y_1,y_2,\dots,y_L\}$ and image $\III$, the inner loop loss now adds a value $\gamma_t$ for each time-step $t$:

\begin{equation} \label{eq:inner-loop-loss}
    \mathcal{L}^{inner} = - \frac{1}{L} \sum_{t=1}^{L} \gamma_t \log p(y_t|\III; \theta),
\end{equation}

which is a modified version of cross-entropy, including $\gamma_t$ values inside the summation. In order to calculate $\gamma_t$, gradient information from the final classification layer is used. The idea is that the gradients provide information related to disagreement, i.e., what knowledge is missing from the model that still needs to be learned. Specifically, let the weights of the final classification be denoted as $\phi$. The gradients of the $t$'th instance loss with respect to the weights of the final classification layer are used, denoted as $\nabla_{\phi} \mathcal{L}^t$, in combination with the gradients of the mean loss (Eq.~\ref{eq:cross-entropy-loss}), denoted as $\nabla_{\phi} \mathcal{L}$. Both inputs are concatenated and fed as input to a network $g_{\psi}$, leading to character instance-specific weight $\gamma_t$, where
$
    \gamma_t = g_{\psi}( [\nabla_{\phi}\mathcal{L}^{t}; \nabla_{\phi}\mathcal{L}] )
$.
$g_{\psi}$ takes the form of a 3-layer MLP with parameters $\psi$, followed by a sigmoid layer to produce a scalar output value in the range [0, 1].

\paragraph{Learnable layer-wise learning rates:} \label{sec:llr}

The inner loop learning rate used in MAML is replaced by a learnable one~\citep{li2017metasgd}. Specifying a learnable learning rate for every model parameter allows the model to express differences between what parameters should be updated more or less. However, using a learning rate for every parameter also doubles the parameter count, which is prohibitive. Therefore, learning rates are used for individual layers in the model, which are trained along with all the other parameters. This is also shown in Algorithm~\ref{alg:metahtr-train}.

\begin{algorithm}[tb]
    \caption{Training for MetaHTR, adapted from~\citet{bhunia2021metahtr}.}
    \label{alg:metahtr-train}
    \begin{algorithmic}[1]
    \Require Training dataset $\DDD = \set{\DDD_1, \DDD_2, \dots,
                \DDD_{|\WWW^{tr}|}}$
    \Require $\beta$: learning rate
    \State Initialize $\theta, \psi, \alpha$
    \While {not done}
        \State Sample writer-specific $\TTT_i = \set{D^{tr}_i, D^{val}_i} \sim p(\TTT)$
        \ForAll {$\TTT_i$}
            \State Evaluate inner objective: $\LLL^{inner}(\theta; D^{tr}_i)$
            \State Adapt: 
    $\theta_i' = \theta - \alpha \nabla_{\theta} \mathcal{L}^{inner}(\theta; D_i^{tr})$
            \State Compute outer objective: $\LLL^{outer}(\theta'_i; D^{val}_i)$
        \EndFor
        \State Update meta-parameters: $(\theta, \psi, \alpha) \leftarrow
        (\theta, \psi, \alpha) - \beta \nabla_{(\theta, \psi, \alpha)} \sum_{\TTT_i
        }\mathcal{L}^{outer}(\theta_i'; D^{val})$ \EndWhile
    \end{algorithmic}
\end{algorithm}

\begin{algorithm}[tb]
    \caption{Inference for MetaHTR, adapted from~\citet{bhunia2021metahtr}.}
    \label{alg:metahtr-inference}
    \begin{algorithmic}[1]
    \Require Testing dataset $\DDD = \set{\DDD_1, \DDD_2, \dots, \DDD_{|\WWW^{test}|}}$
    \Require Meta-learned model parameters $\set{\theta, \psi, \alpha}$
    \Require A given writer $j$
        \State Evaluate inner objective: $\LLL^{inner}(\theta; D^{tr}_j)$
        \State Adapt: $\theta_j' = \theta - \alpha \nabla_{\theta} \mathcal{L}^{inner}(\theta; D_j^{tr})$
    \end{algorithmic}
    \Return \textit{Writer-specialized} HTR model parameters $\theta'_j$
\end{algorithm}

\subsubsection{Meta-learning evaluation}

We evaluate several variants of the MAML/MetaHTR approach. One downside of the MAML approach and MetaHTR, in particular, is that it leads to a notable increase in memory and computational requirements. We, therefore, analyze variations of the MAML-based approach to investigate to what degree it can be simplified. Concretely, we experiment with three different models: MAML, MAML + llr, and MetaHTR.

\begin{enumerate}
    \item \textbf{MAML:} The original MAML algorithm, as proposed in~\citet{finn2017maml}, using the sequence-based cross-entropy loss function shown in~\ref{eq:cross-entropy-loss}.
    \item \textbf{MAML + llr:} The MAML algorithm is complemented with learnable inner loop learning rates (Section~\ref{sec:llr}). This alleviates the need to manually set the inner loop learning rate, at the cost of only a few hundred additional parameters (see Appendix~\ref{tab:extra-params-meta})
    \item \textbf{MetaHTR:} The full MetaHTR model is explained in Section~\ref{sec:metahtr}. A downside of the MetaHTR approach is the additional complexity that it introduces. Next to the calculation of higher-order derivatives as part of the MAML algorithm, MetaHTR also requires an additional backward pass in order to calculate the instance-specific weights. This makes the approach expensive both in terms of computation and in terms of memory usage, therefore making it challenging to scale to larger contexts such as sentence-level HTR. 
\end{enumerate}

%%%%%%%%%%%%%%%%%%%%%%%%%%%%%%%%%%%%%%%%%%%%%%%%%%%%%%%%%%%%%%%%%%%%%%%%%%%%%%
\subsection{Writer codes} \label{sec:writer-code-approach}
Our second attempt to include writer information into the base HTR models is based on the idea of representing style or writer information as a compact feature vector. In speech recognition, such a code is known as a \textit{speaker code}~\citep{abdel2013speakeradapt3}. We take a similar approach by trying to model writers or styles using a small feature vector, which is used to adapt the weights of an existing HTR model. We will refer to such vectors as \textit{writer codes}. A writer code is a dense feature vector $\mathbf{x} \in \RR^M$, where $M$ is set based on the desired representational capacity. A relevant property of writer codes is that they should be able to obtain them even for writers that are not part of the initial training set. Writer codes have certain properties that make them appealing as a method for writer-adaptive HTR: they are efficient to compute and often require minimal changes to a base architecture.

\subsubsection{Code insertion} \label{sec:code-insertion}

First, we address the question of how the codes should be inserted into the base model for effective adaptation. A comprehensive evaluation of possible methods for code insertion is beyond the scope of this paper, but we note here that, based on various experiments, naive insertion of codes into the base models can easily deteriorate base-level performance. Notably, naively modifying batch normalization (batch norm) parameters can lead to catastrophic forgetting. Furthermore, we found that adapting only certain key layers of the network, such as the last layers of the ResNet backbone, was not sufficient to allow for effective adaptation. Instead, an effective form of vector-based adaptation comes from fine-tuning the normalization layers of the model. This approach is inspired by work on generative models, such as conditional GANs~\citep{karras2019stylegan,zhang2022divergan} and methods for style transfer~\citep{dumoulin2016learned, ulyanov2016adain}. Previous work in the field of style transfer suggests that in order to adapt features to a particular style, it can be sufficient to specialize scaling and shifting parameters after normalization layers, conditioned on style information~\citep{dumoulin2016learned}. We adopt a similar approach, where we update the learnable weights of the normalization layers in our network, conditioned on a specific writer code. Specifically, we focus on batch normalization layers, which are present in the ResNet backbone\footnote{It is worth noting that for the FPHTR model, layer normalization is used in addition to batch normalization. However, we found no concrete benefit in adjusting these normalization layers.}. Given a minibatch of activations $B = \set{x_{1,\dots,m}}$, batch normalization layers are of the following form:

\begin{equation} \label{eq:batchnorm}
    y_i = \frac{x_i - \mu_B}{\sqrt{\sigma^2_B + \epsilon}} \cdot \gamma + \beta,
\end{equation}

where $\gamma$ and $\beta$ are learnable parameter vectors of size equal to the number of channels in the input. The $\epsilon$ parameter is a small constant added for numerical stability. The normalization statistics are calculated along the batch dimension:

\begin{equation}
    \mu_B = \frac{1}{m} \sum_{i=1}^{m} x_i, \quad \quad
    \sigma^2_B = \frac{1}{m} \sum_{i=1}^{m} (x_i - \mu_B)^2.
\end{equation}

For inserting writer codes into the neural network, we modify the $\beta$ and $\gamma$ parameters based on an input code (corresponding to an approach called \textit{conditional batch normalization}~\citep{devries2017condbn}). Given pretrained parameters $\beta_c$ and $\gamma_c$, changes in these parameters are predicted based on an input code  $e$ and a two-layer MLP:

\begin{equation}
    \Delta \beta = \phi_1(e), \quad \quad \Delta \gamma = \phi_2(e),
\end{equation}

where $\phi_1$ and $\phi_2$ are MLPs. The predicted deltas are then added to the original $\beta_c$ and $\gamma_c$ parameters:
$
    \hat{\beta}_c = \beta_c + \Delta \beta_c, \hat{\gamma}_c = \gamma_c + \Delta \gamma_c
$,
where $\hat{\beta}_c$ and $\hat{\gamma}_c$ replace the batch norm parameters for the current forward pass. All other parameters are frozen during training, including $\beta$ and $\gamma$. By changing the $\gamma$ and $\beta$ affine parameters that follow normalization, there is great flexibility in changing the intermediate feature maps according to the specifics of a particular code, while the risk of catastrophic forgetting is mitigated by keeping the original batch normalization weights largely intact.

\subsubsection{Code creation} \label{sec:learned-codes}

Given the conditional batch normalization method for inserting writer codes into an HTR model, we turn to the question of how we create writer codes. An important criterion is that writer codes are not created under a closed writer set assumption; we should be able to instantiate them for novel writers as well. We experiment with two kinds of writer codes: learned codes, and codes based on statistical writer information (Hinge codes and style codes).

\paragraph{Learned codes:}

Learned writer codes are obtained by training them in the same way as the weights of the network. A similar idea is commonly seen in NLP (e.g., \citet{devlin2018bert}), where for each token in a predefined vocabulary, an associated vector representation is learned (often called an ``embedding'') that is more expressive than a one-hot vector indicating the identity of the token. Note that this approach implies a fixed set of writer codes initialized at the start of training -- one for each writer in the training set. In the case when a new writer is presented that is unseen during training, we follow~\citet{abdel2013speakeradapt3} by randomly initializing a new writer code, followed by one or several gradient steps on the newly initialized code, using a small batch of labeled writer-specific data. 

\paragraph{Hinge codes:}

When it comes to capturing writer individuality, there exists a rich literature on this topic in the field of writer identification~\citep{schomaker2007advances}. In contrast to the learned features discussed in the previous section, features for writer identification are often handcrafted or statistical in nature. One of the more successful features for writer identification is the Hinge feature~\citep{bulacu2007hinge}, which uses a probability distribution of the angle combination of two hinged edge fragments to characterize writer individuality. The assumption here is that these features can lead to a meaningful clustering of writers based on their style differences. These writer codes are attractive because they are easy to calculate and do not require additional adaptation data at inference time.

\paragraph{Style codes:}

We also focus on generic style clusters in feature space, rather than features that are highly writer-specific. For example, style clusters could point to high-level writing styles such as cursive or mixed cursive. We perform k-means clustering on Hinge codes to obtain generic style clusters. For each style cluster, we train a writer code using backpropagation. Thus, given an image input, we find the closest style cluster based on the Hinge features and map the style cluster identity to a learned writer code that is updated using gradient descent. 

%%%%%%%%%%%%%%%%%%%%%%%%%%%%%%%%%%%%%%%%%%%%%%%%%%%%%%%%%%%%%%%%%%%%%%%%%%%%%%%%%%%
%%%%%%%%%%%%%%%%%%%%%%%%%%%%%%%%%%%%%%%%%%%%%%%%%%%%%%%%%%%%%%%%%%%%%%%%%%%%%%%%%%%
\section{Experiments} \label{chap:experiments}

\subsection{Dataset} \label{sec:dataset}
\begin{figure}[!htb]
    \centering
    \begin{subfigure}{0.3\textwidth}
        \includegraphics[width=0.8\textwidth]{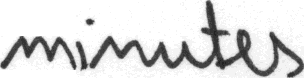}
    \end{subfigure}
    \begin{subfigure}{0.3\textwidth}
        \includegraphics[width=0.8\textwidth]{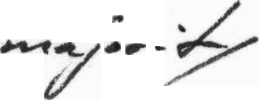}
    \end{subfigure}
    \begin{subfigure}{0.3\textwidth}
        \includegraphics[width=0.8\textwidth]{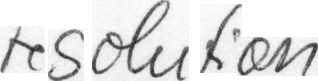}
    \end{subfigure}
    \caption{Examples of word images from the IAM dataset.}
    \label{fig:iam_word_example}
\end{figure}

We use the IAM dataset~\citep{marti2002iam} for evaluation, using word-level images. The dataset consists of English handwritten texts contributed by 657 writers, making a total of 1,539 handwritten pages consisting of 115,320 segmented words. The data is labeled at the sentence, line, and word level. Examples of word images are shown in Fig.~\ref{fig:iam_word_example}. For splitting the data into a training, validation, and test set, we use the widely used Aachen splits~\citep{aachensplits}. An important property of these splits is that the writer sets are disjoint, i.e., writers seen during training are not seen during testing. The Aachen splits contain 500 writers making up a total of 75,476 images.

%%%%%%%%%%%%%%%%%%%%%%%%%%%%%%%%%%%%%%%%%%%%%%%%%%%%%%%%%%%%%%%%%%%%
\subsection{Implementation details}

\paragraph{Base models:}

Character error rate (CER) and word error rate (WER) are used for evaluation, with the best model chosen based on the lowest WER. We use a character-level vocabulary, converting all characters to lowercase. No linguistic post-processing on word predictions is used. We report average performance over five random seeds, along with standard deviations for all results. For training of the base models, the Adam optimizer is used~\cite{kingma2014adam}, with $\beta_1 = 0.9$ and $\beta_2 = 0.999$. We use gradient clipping to avoid exploding gradients based on the L2-norm of the gradient vector. All models are implemented using PyTorch~\cite{pytorch}, using a single Nvidia V100 GPU with 32GB of memory. See appendix Table~\ref{tab:base-hyperparams} for full details about hyperparameter settings. We use random image rotation, scaling, brightness, contrast adjustment, and Gaussian noise to increase image diversity. We reduce the resolution by 50\% to reduce memory footprint while keeping the text legible.

\paragraph{Meta-learning:}

Given the $K$-shot $N$-way meta-learning formulation, we use $K = 16$ and $N = 8$, following~\citet{bhunia2021metahtr}. This means that during adaptation, a batch of $K = 16$ writer-specific examples are used to adapt the model to a specific writer, and outer loop gradients are averaged over $N = 8$ writers (see Eq.~\ref{eq:maml-outer-loop}). During training, we randomly sample writer-specific batches of size $2K$, split into a support and query set of size $K$. At test time, we use all examples for a given writer: Given the $j$'th writer with $N_j$ total examples, we randomly split the data into a support batch (adaptation batch) of size $K$, and use the remaining $N_j - K$ examples for evaluation of the adapted model. Performance per writer is averaged over ten runs. For all models, we use dropout in the outer loop. Batch norm statistics are fixed to their running values and not updated during training, as this led to more stable performance (see Appendix~\ref{appendix:batch-normalization-in-maml} for a more extensive discussion concerning the particulars of using batch normalization in combination with MAML). We use the learn2learn library~\citep{learn2learn} for implementing all meta-learning methods. Full hyperparameter settings are shown in the Appendix (Table~\ref{tab:hyperparameters-meta}).

\paragraph{Writer codes:}

For the learned writer codes discussed in Section~\ref{sec:learned-codes}, we require adaptation data at test time to initialize codes for novel writers. Splitting of writer data is done in the same way as for meta-learning. During training, the weights of the trained HTR model are frozen, and only the writer code values and the parameters of the conditional batch norm MLPs are updated. We use a code size of 64 and an adaptation batch size of 16. For style codes, we use k-means clustering with $k = 3$, based on validation set performance. Complete hyperparameters are shown in Table~\ref{tab:hyperparameters-writer-code} in the Appendix.

%%%%%%%%%%%%%%%%%%%%%%%%%%%%%%%%%%%%%%%%%%%%%%%%%%%%%%%%%%%%%%%%%%%%%%%%%%%%%%%%%%%
%%%%%%%%%%%%%%%%%%%%%%%%%%%%%%%%%%%%%%%%%%%%%%%%%%%%%%%%%%%%%%%%%%%%%%%%%%%%%%%%%%%
\section{Results} \label{chap:results}

\subsection{Base models}

The results for the base models on the IAM validation and test set are shown in Table \ref{tab:base-results}. We report average performance as well as the performance of the best run. From the results in Table~\ref{tab:base-results}, we can see that the Transformer-based model (FPHTR) outperforms the LSTM-based model (SAR) on validation and test, both for the smaller 18-layer case (15-18M weights) and the larger 31-layer case (52-58M weights). This difference is significant in the case of the larger 31-layer models, with FPHTR outperforming SAR on the test with a difference of 4.1 WER and 4.8 CER. For the smaller 18-layer models, FPHTR outperforms SAR by a difference of 0.5 WER and 0.7 CER.

\begin{table}[!tbh]
    \centering
    \caption{Results of the base models on the IAM val and test set (lower is better).}
    \begin{tabular}{lcccccccc}\toprule
         & \multicolumn{4}{c}{\bf Val} & \multicolumn{4}{c}{\bf Test} \\
         \cmidrule(lr){2-5} \cmidrule(lr){6-9}
         & \multicolumn{2}{c}{\bf WER} & \multicolumn{2}{c}{\bf CER} & \multicolumn{2}{c}{\bf WER} & \multicolumn{2}{c}{\bf CER} \\
         \cmidrule(lr){2-3} \cmidrule(lr){4-5} \cmidrule(lr){6-7} \cmidrule(lr){8-9}
                   & Avg. & Best & Avg. & Best & Avg. & Best & Avg. & Best \\\midrule
        SAR-18     & $16.3 \pm 0.6$ & 15.5 & $13.5 \pm 1.0$ & 12.2  & $20.7 \pm 0.8$ &
            19.7 & $17.3 \pm 0.8$ & 15.8 \\
        FPHTR-18   & $\bf{16.0} \pm 0.4$ & 15.3 & $\bf{12.6} \pm 0.4$ & 12.1  & $\bf{20.2} \pm 0.2$ &
            19.9 & $\bf{16.6} \pm 0.3$ & 16.4 \\\midrule
        SAR-31     & $14.9 \pm 0.2$ & 14.7 & $11.3 \pm 0.5$ & 10.6 & $19.7 \pm 0.7$ &
            18.8 & $15.7 \pm 1.0$ & 14.5 \\
        FPHTR-31   & $\bf{11.6} \pm 0.3$ & 11.1 & $\bf{7.9} \pm 0.4$ & 7.5 & $\bf{15.6} \pm 0.8$ &
            14.6 & $\bf{10.9} \pm 0.7$ & 10.0  \\\bottomrule
    \end{tabular}
    \label{tab:base-results}
\end{table}

%%%%%%%%%%%%%%%%%%%%%%%%%%%%%%%%%%%%%%%%%%%%%%%%%%%%%%%%%%%%%%%%%%%%%%%%%%%%%%%%%%%%%%
\subsection{Meta-learning} \label{sec:meta-learning-results}
Results for meta-learning are shown in Table~\ref{tab:meta-learning-res}. It should be noted that since all models presented here make use of additional adaptation data at test time, a direct comparison between the base models in Table~\ref{tab:base-results} is not directly meaningful. In other words, the MAML-based models have access to parts of the test data as part of their adaptation procedure. Therefore, we devise a different baseline, by evaluating the base models after performing fine-tuning on the same adaptation data that is made available to the MAML-based models. Specifically, we fine-tune the final classification layer of a base model using the adaptation data. We use the Adam~\citep{kingma2014adam} optimizer with a learning rate of 1e-3 for 3 optimization steps. Due to persistent out-of-memory errors for the SAR-31 MetaHTR model\footnote{Another performance-related issue worth mentioning is that MetaHTR requires calculation of instance-specific gradients, which, at the time of running the experiments, is something that is not supported in batch form in the PyTorch library. Therefore, this required a manual calculation of instance-specific gradients using a for-loop, which made the MetaHTR training procedure considerably slower than MAML. This problem is something that can be fixed using additional software, but the additional complexity of MetaHTR due to the extra backward pass remains.}, we only include FPHTR-31 in addition to the smaller 18-layer variants.  From these results, we can see that MetaHTR performs best, improving upon the baseline by 1.4, 2.0, and 1.7 WER for FPHTR-18, SAR-18, and FPHTR-31, respectively.

\begin{figure}
    \centering
    \includegraphics[trim=60 10 70 55, clip, width=0.9\textwidth]{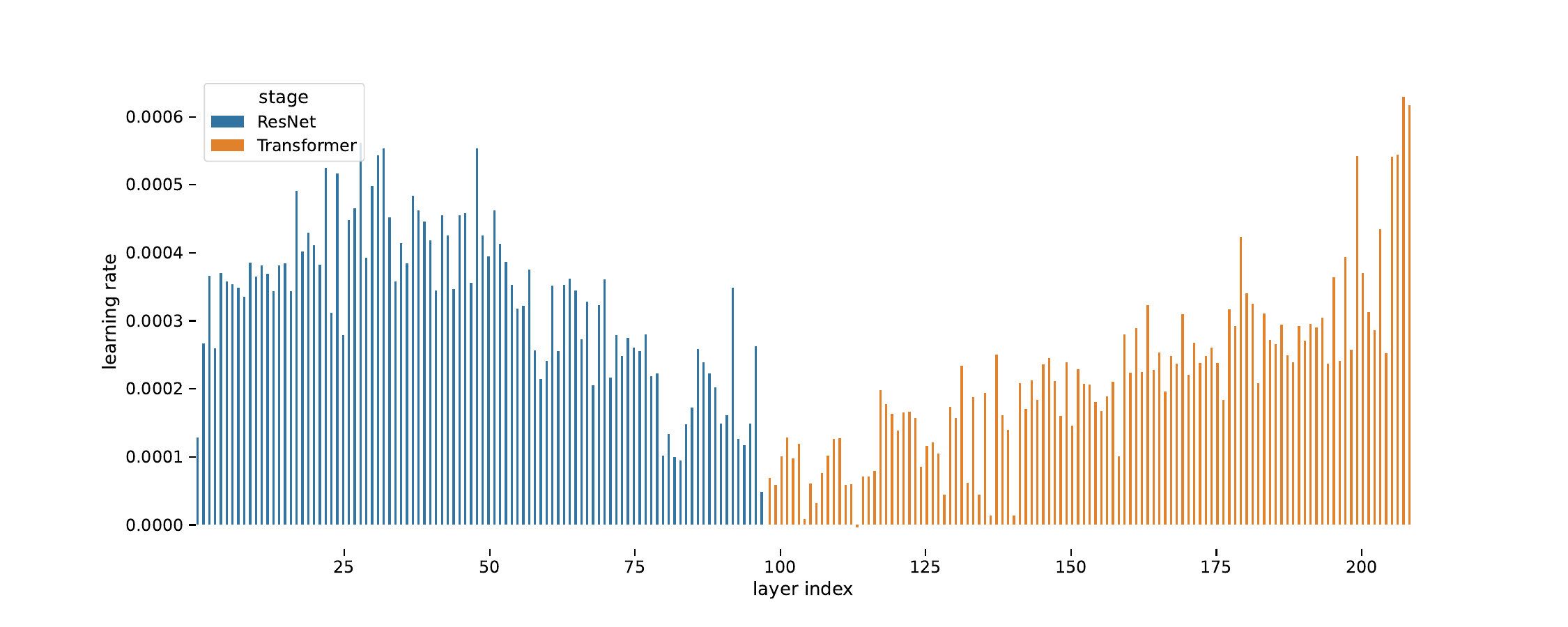}
    \caption{Learned per-layer learning rates for the MAML + llr model, for FPHTR-31.}
    \label{fig:lr-per-layer-fphtr31}
\end{figure}

We plot the learned inner loop learning rates in Fig.~\ref{fig:lr-per-layer-fphtr31}, to get an idea of the relative weight assigned to each layer in the adaptation process. We show learned inner loop learning rates for two randomly chosen runs of the FPHTR-18 and FPHTR-31 models using MAML + llr (we include the figure for FPHTR-18 in the appendix, Fig.~\ref{fig:lr-per-layer}). Looking at these plots, we see a relatively high weight assigned to the ResNet layers, decreasing towards the head of the network. For the Transformer module, we observe an increasing trend in the learning rates across layers. This is an indication that the lower layers of the Transformer network require relatively fewer adaptation than layers closer to the output, with the final classification layer requiring the most adaptation.

It is worth noting here that the performance improvements for MetaHTR (between 1.4 to 2.0 WER compared to the baseline) are much smaller than reported in the original paper~\citep{bhunia2021metahtr}, where MetaHTR improved upon the SAR baseline by a difference of 7.1 WER, and 6.8 after naive fine-tuning on the adaptation data. In email correspondence with the authors of the MetaHTR paper, we were not able to resolve the cause of this discrepancy. Furthermore, due to the lack of published code by the MetaHTR authors, it is difficult to cross-verify the MetaHTR results.

\begin{table}[!htb]
    \caption{Meta-learning results on the IAM test set, measured in WER (lower is better).}
    \centering
    \begin{tabular}{cccc}
        \toprule
                   & FPHTR-18 & SAR-18 & FPHTR-31 \\\midrule
        Baseline   & $20.0 \pm 0.2$ & $20.6 \pm 0.6$ & $15.3 \pm 0.7$ \\\midrule
        MAML       & $19.1 \pm 0.3$ & $19.5 \pm 0.7$ & $14.3 \pm 0.3$ \\
        MAML + llr & $19.3 \pm 0.5$ &  $19.3 \pm 0.7$ & $14.3 \pm 0.2$ \\
        MetaHTR    & $\bf 18.6 \pm 0.4$ & $\bf 18.6 \pm 0.5$ & $\bf 13.5 \pm 0.2$ \\
        \bottomrule
    \end{tabular}
    \label{tab:meta-learning-res}
\end{table}

%%%%%%%%%%%%%%%%%%%%%%%%%%%%%%%%%%%%%%%%%%%%%%%%%%%%%%%%%%%%%%%%%%%%%%%%%%%%%%%%%%%
\subsubsection{Testing the adaptation premise of MetaHTR} \label{sec:testing-adaptation-premise}

An important question concerning the efficacy of MetaHTR is to what degree it truly \textit{adapts} based on a set of writer-specific images at test time. This is an important premise, since the additional computational overhead of MetaHTR as well as the increased complexity compared to regular neural network training is supposedly warranted by a clear goal: An ability to adapt in a flexible way to various writers leading to a performance improvement compared to a writer-unaware model. In the words of the authors, the goal of MetaHTR is to offer a ``adapt to my writing button''~\citep{bhunia2021metahtr}, where one is asked to write a specific sentence in order to make recognition performance of that handwriting more accurate.

Note that because the MetaHTR objective function and training procedure are different from the training procedure used for the baseline, it is not clear that the improved performance of MetaHTR is due to writer adaptation. The MetaHTR objective function is designed for writer-specific adaptation, but it may simply be a more effective way to train the neural network, regardless of whether writer adaptation is performed or not. The writer adaptation performed at test time is what is supposed to make MetaHTR writer adaptive. Therefore, if it is writer adaptive, it should perform better than MetaHTR \textit{without} writer adaptation at test time. In order to test this, we leave out the writer-specific adaptation. More concretely, we train MetaHTR the same way as done before but evaluate it without performing inner loop adaptation on a support batch of $K$ images. Results are shown in Table~\ref{tab:meta-learning-res-adaptation}.  The additional benefit of adaptation is 0.2 WER for FPHTR-18, 0.7 WER for SAR-18, and 0.7 WER for FPHTR-31.  We use a two-sample t-test to measure the statistical significance of the difference in results.  Using a significance level $\alpha = 0.05$, we observe that the difference in results is not significant for FPHTR-18 ($p = 0.4143$) and SAR-18 ($p = 0.0832$), but \emph{is} significant for FPHTR-31 ($p = 0.0001$). In other words, adaptation only shows a significant effect for the larger FPHTR-31 model, but not for the smaller 18-layer variants.

\begin{table}[!htb]
    \caption{MetaHTR performance with and without writer adaptation, measured in WER.}
    \centering
    \begin{tabular}{cccc}
        \toprule
                          & FPHTR-18 & SAR-18 & FPHTR-31 \\\midrule
        w/ adaptation     & $18.6 \pm 0.4$ & $18.6 \pm 0.5$ & $13.5 \pm 0.2$ \\
        w/o adaptation    & $18.8 \pm 0.4$ & $19.3 \pm 0.5$ & $14.2 \pm 0.2$ \\
        \bottomrule
    \end{tabular}
    \label{tab:meta-learning-res-adaptation}
\end{table}

%%%%%%%%%%%%%%%%%%%%%%%%%%%%%%%%%%%%%%%%%%%%%%%%%%%%%%%%%%%%%%%%%%%%%%%%%%%%%%%%%%%%%%
\subsection{Writer codes}  \label{sec:writer-code-results}
We show results for all writer codes in Table~\ref{tab:results-writer-code}. From the table, it can be seen that the learned codes do not improve upon the performance of the baseline. The fact that writer codes at test time are created by random initialization followed by only a small number of gradient steps is a potential factor here -- codes trained in this way seem to hurt performance rather than improve it.

Next, we consider Hinge and style codes. Both methods outperform the baseline. For the Hinge code, this is a difference of 1.7 and 1.6 WER for FPHTR and SAR, respectively.  A similar performance improvement can be seen for the style code, obtained by clustering Hinge codes with a single learned code per style cluster.  In this case, the difference is 1.8 and 1.7 WER for FPHTR and SAR, respectively. 

Although these results show improvement compared to the baselines, they do not provide adequate insight into the efficacy of the codes themselves.  Recall from Section~\ref{sec:code-insertion} that conditional batch normalization uses a 3-layer MLP with the writer codes as input to predict changes to the original batch norm weights.  It is possible that the MLP learns effective bias vectors that improve performance regardless of the writer code input, i.e., the writer code could simply be ignored (e.g., assigned zero weights). To test this, we replace the writer codes with a zero code that contains no writer information whatsoever, i.e., a vector with only zero values. As seen from Table~\ref{tab:results-writer-code}, this leads to almost identical performance compared to both the Hinge and style code.  This is a strong indication that writer information is not the direct cause of the increase in performance, but rather, \textit{conditional batch normalization seems to be an effective way to fine-tune the batch norm weights, even without the presence of conditional information}. Although this may be an interesting way to perform general fine-tuning, it does not rely on writer-specific information to make it possible.

\begin{table}[!htb]
    \centering
    \caption{Writer code results on the IAM test set, measured in WER (lower is better).}
    \begin{tabular}{lcc}\toprule
                      & FPHTR-18 & SAR-18 \\\midrule
        Baseline      & $20.2 \pm 0.2$ & $20.7 \pm 0.8$  \\\midrule
        Learned code  & $24.5 \pm 0.3$ & $23.7 \pm 0.4$  \\
        Hinge code    & $\bf 18.5 \pm 0.2$ & $\bf 19.1 \pm 0.6$  \\
        Style code    & $\bf 18.4 \pm 0.2$ & $\bf 19.0 \pm 0.6$ \\
        Zero code     & $\bf 18.5 \pm 0.3$ & $\bf 19.0 \pm 0.5$  \\\bottomrule
    \end{tabular}
    \label{tab:results-writer-code}
\end{table}

%%%%%%%%%%%%%%%%%%%%%%%%%%%%%%%%%%%%%%%%%%%%%%%%%%%%%%%%%%%%%%%%%%%%%%%%%%%%%%%%%%%
%%%%%%%%%%%%%%%%%%%%%%%%%%%%%%%%%%%%%%%%%%%%%%%%%%%%%%%%%%%%%%%%%%%%%%%%%%%%%%%%%%%
\section{Discussions} \label{chap:discussion-and-conclusion}

\subsection{Meta-learning}
An appealing aspect of the meta-learning approach is that there is a great deal of flexibility in the way the model can adapt to a writer by differentially updating the layers of the model (e.g., as demonstrated in Fig.~\ref{fig:lr-per-layer-fphtr31}). Nevertheless, the added benefit of writer adaptation using MetaHTR is not obvious, as shown in Section~\ref{sec:testing-adaptation-premise}. Even without using any adaptation data at test time, the MetaHTR model still improves upon the baseline performance. This indicates that more effective representations play a role in the additional performance gains, rather than rapid adaptability of the model parameters, a phenomenon observed before in the literature on meta-learning~\citep{raghu2019anil}. This makes MetaHTR interesting for improving overall model performance, but not necessarily for writer-specific adaptation. Another downside of the MetaHTR approach is the additional complexity that it introduces. Next to the calculation of higher-order derivatives as part of the MAML algorithm, MetaHTR requires an additional backward pass to calculate the instance-specific weights (Section~\ref{sec:instance-weights}). This makes the approach expensive both in terms of computation and memory usage and makes it challenging to scale to larger contexts such as sentence-level HTR. This is exemplified by the fact that we were not able to train MetaHTR in combination with the SAR-31 base model on a 32GB GPU due to persistent out-of-memory errors. This is somewhat problematic given our finding that a deeper model lends itself better to adaptation using MetaHTR than a shallower one. Another example of additional complexity is the difficulty caused by the interaction of MAML with batch normalization (see Appendix~\ref{appendix:batch-normalization-in-maml} for a more extensive discussion on this topic).

Moreover, training of MetaHTR requires a good deal of fine-tuning of various hyperparameters to make it work well, which is something that has also been observed for MAML more broadly~\citep{antoniou2018mamlplusplus}. Given the modest benefits for writer adaptation (0.7 WER in the best case), combined with the increased model complexity, it can be argued that MetaHTR is perhaps not worth the extra investment for writer adaptation. This is especially true given that when more labeled examples are available, a simpler method, such as transfer learning, may be more cost and time effective.

%%%%%%%%%%%%%%%%%%%%%%%%%%%%%%%%%%%%%%%%%%%%%%%%%%%%%%%%%%%%%%%%%%%%%%%%%%%%%%%%%
\subsection{Writer codes}
\begin{figure}
    \centering
    \includegraphics[width=\textwidth]{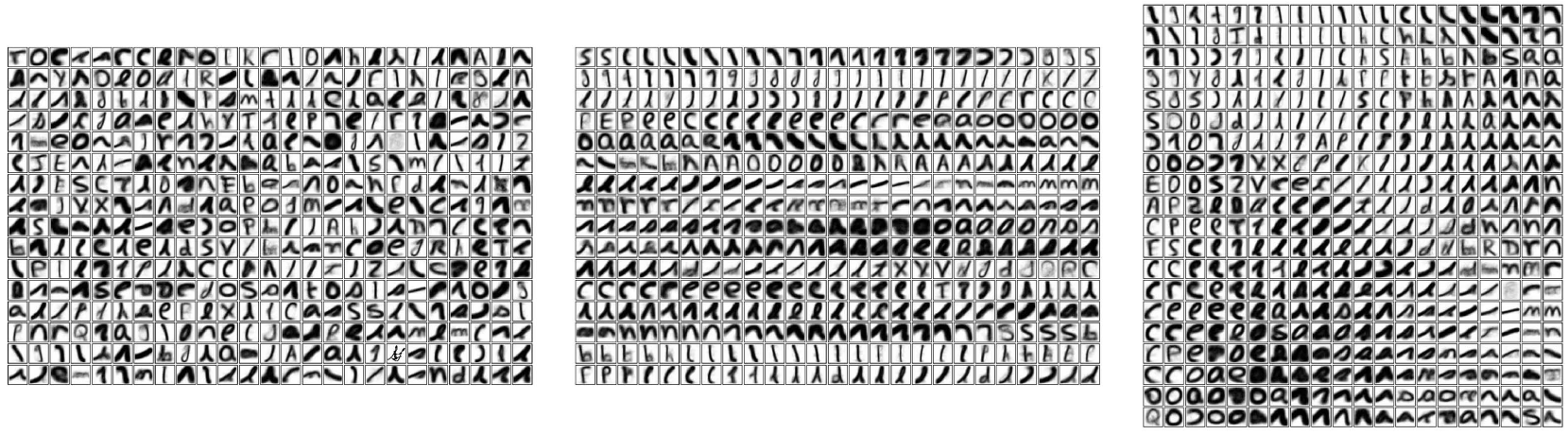}
    \caption{Examples of codebooks that capture shape information based on clustering of
        character shapes. The codebook entries act as prototypes representative of the
        types of shapes commonly seen in handwriting.
        Figure taken from~\citet{bulacu2007hinge}.}
    \label{fig:codebook-example}
\end{figure}

The results in Table~\ref{tab:results-writer-code} show the limited effectiveness of the writer code idea. We showed that statistical features for characterizing writer identity do not show a benefit over a constant zero vector. The fact that the Hinge feature is designed to be independent of the textual context of the handwriting samples may play a role here~\citep{bulacu2007hinge}. An option for future work would be to explore features that lend themselves better to characterize the most relevant writer characteristics, such as idiosyncratic letter shapes that are difficult to classify. For example, a Fraglet approach based on shape codebooks~\citep{bulacu2007hinge} may capture the individual shape features of a particular handwriting more appropriately (see Fig.~\ref{fig:codebook-example}). A histogram can be compiled by matching codebook prototypes with the character shapes observed for an individual writer, counting the matched codebook entries. The normalized histogram can subsequently be used as a vector representation.  

One factor which may play an important role here is data volume. For example, consider automatic speech recognition, where the notion of ``speaker adaptation'' appears to be more common. One facet in which speech and text recognition diverge is the availability of large-scale labeled datasets. Whereas collecting and labeling handwriting samples can be cumbersome and labor-intensive, speech transcriptions are generally easier to obtain.  Thus, if data volume is the critical bottleneck for learning robust representations that lend themselves well to adaptation, methods used in speech recognition relying on large-scale datasets may not transfer as well to HTR. Indeed, as shown by recent work on large language models~\citep{brown2020language}, scale may be a major enabling factor for effective few-shot adaptation.

%%%%%%%%%%%%%%%%%%%%%%%%%%%%%%%%%%%%%%%%%%%%%%%%%%%%%%%%%%%%%%%%%%%%%%%%%%%%%%%%%%%
\section{Conclusion} \label{chap:conclusion}

In this paper, we studied various methods for making neural network-based HTR models writer adaptive. Meta-learning showed the most promising results, with both MAML and MetaHTR leading to improved performance compared to  baseline models. However, we showed that only a relatively small portion of these improvements (between 14-39\%, or 0.2-0.7 WER) can be attributed to writer adaptation, with most of the improvements coming from changes in the way the neural network is trained. It remains to be seen whether MetaHTR could be used to handle more radical domain shifts, as seen, for example, in historical handwriting. Given the observation that writer adaptation using MetaHTR may work better for deeper models, potential future work may focus on scaling up MetaHTR to deeper models. However, memory and/or computational requirements may become prohibitive in this case. Lastly, results show that writer code-based adaptation using learned features or statistical Hinge features does not lead to increased performance. However, updating batch normalization weights may be an effective way to perform general fine-tuning.

%%%%%%%%%%%%%%%%%%%%%%%%%%%%%%%%%%%%%%%%%%%%%%%%%%%%%%%%%%%%%%%%%%%%%%%%%%%%%%%%%%%
%%%%%%%%%%%%%%%%%%%%%%%%%%%%%%%%%%%%%%%%%%%%%%%%%%%%%%%%%%%%%%%%%%%%%%%%%%%%%%%%%%%
\bibliographystyle{unsrtnat}
% \bibliography{refs}

\clearpage
\newpage

%%%%%%%%%%%%%%%%%%%%%%%%%%%%%%%%%%%%%%%%%%%%%%%%%%%%%%%%%%%%%%%%%%%%%%%%%%%%%%%%%%%
%%%%%%%%%%%%%%%%%%%%%%%%%%%%%%%%%%%%%%%%%%%%%%%%%%%%%%%%%%%%%%%%%%%%%%%%%%%%%%%%%%%
\appendix
\section{Batch normalization in MAML} \label{appendix:batch-normalization-in-maml}
In this section, we discuss the role of batch normalization in the MAML-based models. For MAML and MetaHTR models, using batch normalization~\citep{ioffe2015batchnorm} in the right way was generally crucial to obtain good performance, and would often determine whether a model would work at all. Although the current discussion is not directly relevant to the main narrative of the paper, we include it here for the sake of completeness, as it may be useful for future researchers using MAML-based methods.

It has been reported in~\citet{antoniou2018mamlplusplus} that the implementation from the original MAML paper~\citep{finn2017maml} makes use of batch statistics to normalize the activations in batch normalization layers and that~\citet{antoniou2018mamlplusplus} discovered through experimentation that standard batch normalization using stored statistics does not work well. There is a seemingly plausible explanation for why batch normalization could be problematic when training on radically different tasks. During normal neural network training, batches of data are randomly sampled, which, if large enough, have statistics that are close to the dataset statistics. This implies that the batch statistics will remain relatively stable during training. However, introducing task-specific batches of data can potentially lead to large shifts in activation statistics during training, since batches of data are now \textit{task-specific}, i.e., one batch contains a single task. Especially as the number of inner loop optimization steps is increased, the deviation from the global mean and variance will tend to grow. 

Nevertheless, based on our experiments, we found the opposite to hold true for our HTR models. Using batch statistics degraded performance, and depending on the base model, it would lead to consistently inferior performance. Numerous setups have been tried out in this regard, based on what was proposed in~\citet{antoniou2018mamlplusplus}, e.g., fixing the $\gamma$ parameter in the batch normalization layers, or only using batch statistics in the inner loop, but none of these setups yielded good results. 

The explanation for this discrepancy may lie in the nature of the tasks used in MAML. In traditional MAML setups such as few-shot image classification, introducing a new task implies introducing one or several new image classes. The image distribution may therefore change radically, along with the distribution of the intermediate layer activations, and the previously stored statistics may not work well anymore. By contrast, in the HTR setting, different handwriting styles may be similar enough that shared statistics can still be used for normalization.

Notably, the effect of batch normalization was much stronger for the LSTM-based model (SAR). For the SAR base model, using batch statistics for normalization would lead to a significant drop in performance to about 40\% WER. For the FPHTR model, performance was generally worse than with stored statistics, but only by a margin of a few points. Note that the only place where batch normalization takes place is in the ResNet backbone (which FPHTR and SAR both use). Therefore, the LSTM model seems to be more sensitive to the changes in normalization statistics expressed in the ResNet output. Recall that the structure of SAR is such that the output of the ResNet encoder is passed through an initial encoder LSTM processing image strips, followed by a decoder LSTM for language decoding using 2D attention. One difference between the FPHTR and SAR models is that FPHTR uses layer normalization~\citep{ba2016layernorm} following the multi-head attention modules. By contrast, SAR uses no normalization layers after the ResNet encoder. Possibly, this could result in a larger sensitivity to changes in the ResNet output distribution, since the additional variability does not get normalized along the way. 

~\\
\vfill

\clearpage
\newpage

%%%%%%%%%%%%%%%%%%%%%%%%%%%%%%%%%%%%%%%%%%%%%%%%%%%%%%%%%%%%%%%%%%%%%%%%%%%%%%%%%%%
\section{Hyperparameters} \label{appendix:hyperparameters}
In this section, we include all relevant hyperparameters used to train the models in Section~\ref{chap:methods}. We show hyperparameters for the base models in Table~\ref{tab:base-hyperparams}, hyperparameters for writer code models in Table~\ref{tab:hyperparameters-writer-code}, and meta-learning hyperparameters in Table~\ref{tab:hyperparameters-meta}.

\begin{table}[!htb]
    \caption{Hyperparameters for the base HTR models.}
    \centering
    \begin{tabular}{lccc}
         \toprule
                            & \bf FPHTR-\{18,31\} & \bf SAR-18 & \bf SAR-31 \\
         \midrule
         Batch size         & 32  & 32 & 32 \\
         Learning rate      & 1e-4 & 1e-3 & 1e-3 \\
         Adam $\beta_1$     & 0.9  & 0.9  & 0.9 \\
         Adam $\beta_2$     & 0.999  & 0.999  & 0.999 \\
         d\_model           & 260 &  - & -\\
         Feedforward hidden size  & 1024 & - & - \\
         Hidden size LSTM encoder & -  & 256 & 512 \\
         Hidden size LSTM decoder & -  & 256 & 512 \\
         Attention module dim.     & - & 256 & 512 \\
         Dropout encoder    & 0.1  & 0.1 & 0.1 \\
         Dropout decoder    & 0.1  & 0.0 & 0.0 \\
         Transformer heads  & 4  & - & -\\
         Transformer layers   & 6  & - & - \\
         LSTM encoder layers   & - & 2 & 2 \\
         LSTM decoder layers   & - & 2 & 2 \\
         Max sequence length & 55 & 55 & 55 \\
         Max gradient L2-norm & - & 5.0 & 5.0 \\
         \bottomrule
    \end{tabular}
    \label{tab:base-hyperparams}
\end{table}

\begin{table}[!htb]
    \centering
    \caption{Hyperparameters for the writer code approach.}
    \begin{tabular}{lccc}\toprule
                                & Learned code & Hinge code & Style code  \\\midrule
        Learning rate           & 1e-3         & 1e-3       & 1e-3           \\
        Learning rate codes     & 1e-3         & -          & 1e-3          \\
        Batch size              & 128          & 64         & 64             \\
        Code size               & 64           & 465        & 64             \\
        Shots ($K$)             & 16           & -          & -                \\
        Num. clusters ($k$)     & -            & -          & 3               \\
        \bottomrule
    \end{tabular}
    \label{tab:hyperparameters-writer-code}
\end{table}

\begin{table}[!htb]
    \centering
    \caption{Hyperparameters for the meta-learning approach.}
    \begin{tabular}{lcccccc}\toprule
                                & \multicolumn{3}{c}{MAML / MAML + llr} & \multicolumn{3}{c}{MetaHTR} \\
                                \cmidrule(lr){2-4} \cmidrule(lr){5-7}
                                & FPHTR-18 & SAR-18 & FPHTR-31 & FPHTR-18 & SAR-18 & FPHTR-31 \\\midrule
        Learning rate ($\beta$)         & 3e-5 & 1e-4 & 3e-5 & 8e-6 & 1e-4 & 8e-6 \\
        Inner learning rate ($\alpha$)  & 1e-4 & 1e-3 & 1e-4 & - & - & - \\
        MLP ($g_{\psi}$) hidden units   & -   & -   & -   & 128 & 128 & 128 \\
        Shots ($K$)                     & 16  & 16  & 16  & 16  & 16  & 16  \\
        Ways ($N$)                      & 8   & 8   & 8   & 8   & 8   & 8  \\
        Num. inner steps                & 1   & 1   & 1   & 1   & 1   & 1    \\
        Max gradient L2-norm            & 5.0 & 5.0 & 5.0 & 5.0 & 5.0 & 5.0   \\
        \bottomrule
    \end{tabular}
    \label{tab:hyperparameters-meta}
\end{table}

~\\
\vfill

\clearpage
\newpage

%%%%%%%%%%%%%%%%%%%%%%%%%%%%%%%%%%%%%%%%%%%%%%%%%%%%%%%%%%%%%%%%%%%%%%%%%%%%%%%%%%%
\section{Number of parameters per model} \label{appendix:additional-parameters}
We indicate learnable parameter counts for all models below. Base model parameters are shown in Table~\ref{tab:base-parameters}, whereas additional parameters required for each approach in Chapter~\ref{chap:methods} are shown in Tables~\ref{tab:extra-params-writer-code} and \ref{tab:extra-params-meta}.

\begin{table}[!ht]
    \centering
    \caption{Total number of trainable parameters per base model. For each model, the
        total parameter count is decomposed into the constituent submodules.}
    \begin{tabular}{lcc}
        \toprule
                                            & \# parameters \\\midrule
         FPHTR-18                           & \bf 17.8M \\
         \cmidrule(lr){2-2}
         \hspace{5mm}ResNet                 & 11.3M \\
         \hspace{5mm}Transformer decoder    & 6.5M \\\midrule
         SAR-18                             & \bf 14.9M \\
         \cmidrule(lr){2-2}
         \hspace{5mm}ResNet                 & 11.1M \\
         \hspace{5mm}LSTM encoder           & 1.4M \\
         \hspace{5mm}LSTM decoder           & 2.4M \\\midrule
         FPHTR-31                           & \bf 52.6M \\
         \cmidrule(lr){2-2}
         \hspace{5mm}ResNet                 & 46.1M \\
         \hspace{5mm}Transformer decoder    & 6.5M \\\midrule
         SAR-31                             & \bf 57.4M \\
         \cmidrule(lr){2-2}
         \hspace{5mm}ResNet                 & 45.8M \\
         \hspace{5mm}LSTM encoder           & 4.5M \\
         \hspace{5mm}LSTM decoder           & 6.9M \\\bottomrule
    \end{tabular}
    \label{tab:base-parameters}
\end{table}

\begin{table}[!ht]
    \centering
    \caption{Additional number of learnable parameters per writer code variant.}
    \begin{tabular}{lcc}\toprule
                      & FPHTR & SAR \\\midrule
        Learned code  & 1.4M  & 1.4M \\
        Hinge code    & 2.4M  & 2.4M  \\
        Style code    & 1.6M  & 1.6M  \\\bottomrule
        Zero code     & 2.4M  & 1.6M  \\\bottomrule
    \end{tabular}
    \label{tab:extra-params-writer-code}
\end{table}

\begin{table}[!ht]
    \centering
    \caption{Additional number of learnable parameters per meta-learning variant.}
    \begin{tabular}{lccc}\toprule
                    & FPHTR-18 & SAR-18 & FPHTR-31 \\\midrule
        MAML        & 0        & 0      & 0 \\
        MAML + llr  & 173      & 87     & 209 \\
        MetaHTR     & 3.7M     & 14.7M  & 3.7M \\\bottomrule
    \end{tabular}
    \label{tab:extra-params-meta}
\end{table}

~\\
\vfill

\clearpage
\newpage

%%%%%%%%%%%%%%%%%%%%%%%%%%%%%%%%%%%%%%%%%%%%%%%%%%%%%%%%%%%%%%%%%%%%%%%%%%%%%%%%%%%
\section{Additional figures} \label{appendix:additional-figures}
\begin{figure}[htb]
    \centering
    \begin{subfigure}{\textwidth}
        \includegraphics[trim=60 10 70 55, clip, width=\textwidth]{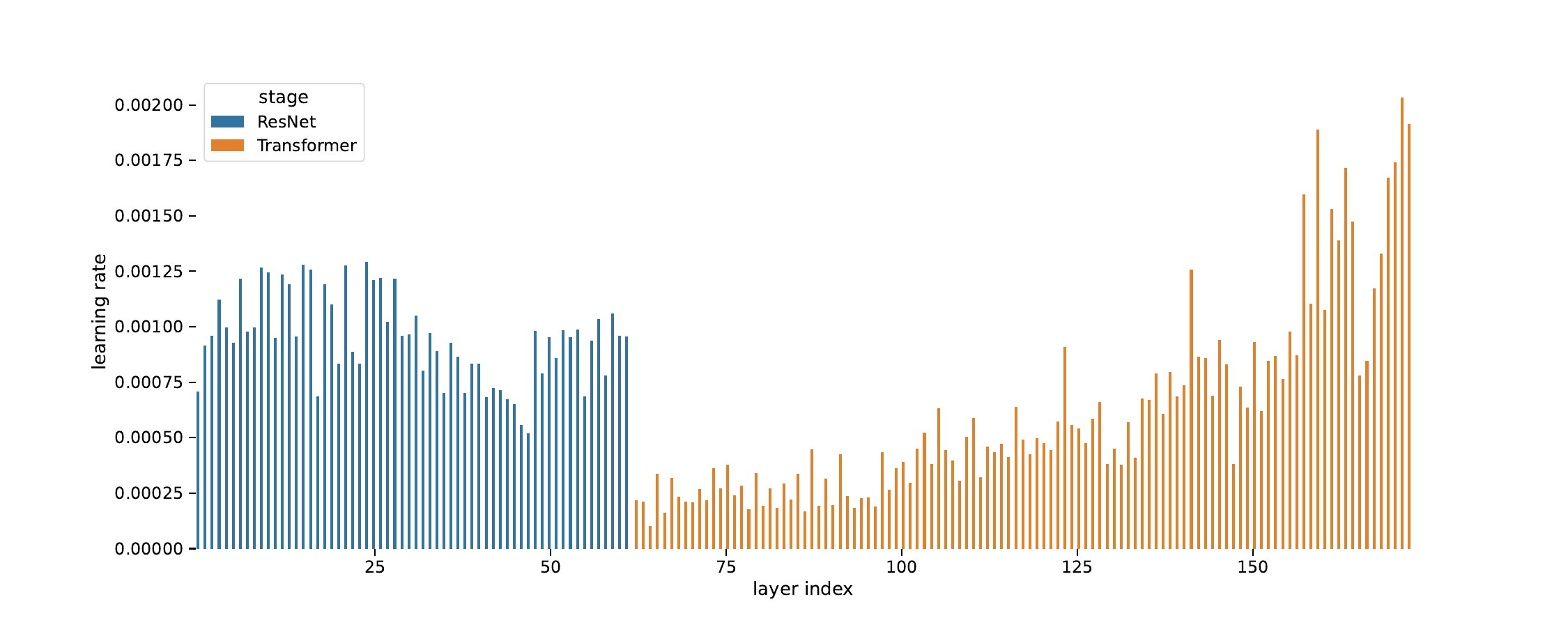}
        \caption{}
    \end{subfigure}

    \begin{subfigure}{\textwidth}
        \includegraphics[trim=60 10 70 55, clip, width=\textwidth]{img/lr-per-layer-fphtr31-seed=1.pdf}
        \caption{}
    \end{subfigure}

    \caption{Learned per-layer learning rates for the MAML + llr model, for (a) FPHTR-18
    and (b) FPHTR-31.}
    \label{fig:lr-per-layer}
\end{figure}

\end{document}